\documentclass{Interspeech2024}[hidelinks]

\interspeechcameraready

\title{Missingness-resilient Video-enhanced Multimodal Disfluency Detection }

\name[affiliation={*1}]{Payal}{Mohapatra}
\name[affiliation={*1}]{Shamika}{Likhite}
\name[affiliation={2}]{Subrata}{Biswas}
\name[affiliation={2}]{Bashima}{Islam}
\name[affiliation={1}]{Qi}{Zhu}

\address{
  $^1$Northwestern University, USA\\
  $^2$Worcester Polytechnic Institute, USA}
\email{(payalmohapatra,shamikalikhite,qzhu)@northwestern.edu, (sbiswas,bislam)@wpi.com}

\usepackage[table,xcdraw]{xcolor}
\usepackage[linesnumbered,ruled,vlined]{algorithm2e}
\usepackage{multirow}
\usepackage{bm}

\usepackage{colortbl}
\usepackage{capt-of}
\usepackage[]{graphicx}
\usepackage{pifont}

\newcommand\blfootnote[1]{%
  \begingroup
  \renewcommand\thefootnote{}\footnote{#1}%
  \addtocounter{footnote}{-1}%
  \endgroup
}

\usepackage{hyperref}

\begin{document}

\maketitle

\begin{abstract}

Most existing speech disfluency detection techniques only rely upon acoustic data. In this work, we present a practical multimodal disfluency detection approach that leverages available video data together with audio. We curate an audio-visual dataset and propose a novel fusion technique with unified weight-sharing modality-agnostic encoders to learn the temporal and semantic context. Our resilient design accommodates real-world scenarios where the video modality may sometimes be missing during inference. We also present alternative fusion strategies when both modalities are assured to be complete. In experiments across five disfluency-detection tasks, our unified multimodal approach significantly outperforms Audio-only unimodal methods, yielding an average absolute improvement of 10\% (i.e., 10 percentage point increase) when both video and audio modalities are always available, and 7\% even when video modality is missing in half of the samples. 



\end{abstract}

\blfootnote{\textsuperscript{$*$} Equal contribution.}

\section{Introduction}
Speech disfluency, encompassing hesitations, repetitions, or prolongations, affects about 1\% of the global population~\cite{yairi2013epidemiology}. Disfluent speech impacts communication, interpersonal skills, social connections~\cite{kirkland2023pardon}, and access to voice-assisted technologies~\cite{clark2020speech}. Thus, advancing research in disfluency is critical for enhancing accessibility and inclusivity in technology~\cite{mohapatra2017novel}.


\begin{figure*}[t]
\centering
\includegraphics[width=0.75\linewidth]{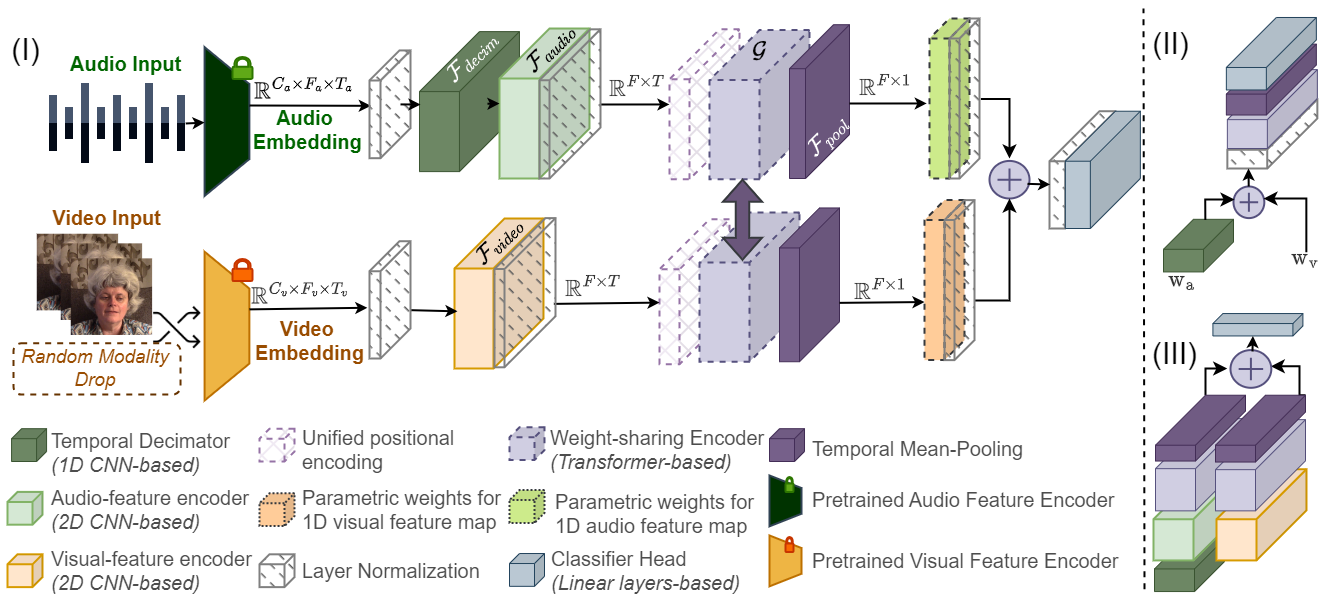}
\vspace{-4pt}
\caption{Illustration of our multimodal learning framework for speech disfluency detection. (I) Unified Modality Fusion Network resilient to missing video modalities, (II) Modality-specific early fusion, and (III) Modality-specific late fusion.}
\vspace{-12pt}
\label{fig:overall_system_diag}
\end{figure*}

Speech disfluency research particularly faces the challenge of lack of public datasets for benchmarking due to high annotation cost, the clinical nature of the task, and the usage of proprietary datasets~\cite{shonibare2022enhancing, TedKPrivate_data}. Certain studies tackle this scarcity by artificially introducing disfluencies into typical speech~\cite{kourkounakis2021fluentnet}. However, this method may oversimplify the nuances of naturally occurring disfluency and might not effectively translate to real-world contexts~\cite{sheikh2022machine}. UCLASS~\cite{uclass}, TORGO~\cite{torgo}, and FluencyBank~\cite{ratner2018fluency} are some of the disfluency-specific speech corpora collected from monologues and interviews with people who stutter ranging from children to dysarthric patients, but they lack labels for disfluencies. Past researchers have conducted private annotations~\cite{TedKPrivate_data, esmaili2017automatic} for these databanks, adhering to an inconsistent taxonomy in the process. To mitigate these shortcomings, Lea et. al~\cite{lea2021sep} have released an annotated dataset, SEP28k, and also provided labels for FluencyBank audio clips with five different classes of disfluencies. All these datasets primarily consist of unimodal audio data, with a few works exploring manual text transcriptions as an additional modality~\cite{romana2023toward,at-disflnt}.

Most existing works in disfluency detection also  heavily rely on audio data, 
employing handcrafted features~\cite{lea2021sep, sheikh2021stutternet}, pretrained language model embeddings~\cite{mohapatra2022speech}, or custom disfluency foundation models with unlabeled atypical speech~\cite{mohapatra2023efficient}. These are often paired with classifiers like statistical models~\cite{chee2009mfcc}, support vector machines~\cite{sheikh2022machine}, or deep neural networks~\cite{lea2021sep,mohapatra2022speech, mohapatra2023efficient, sheikh2021stutternet}. Some recent works incorporate manual text transcriptions~\cite{rocholl2021disfluency} as an additional modality or use off-the-shelf foundation models for untranscribed audio~\cite{at-disflnt, romana2023toward} to improve disfluency detection. Despite outperforming the audio-only features, audio-text multimodal studies recognize the limited expressiveness of disfluency markers through text and the inherent errors in transcribing disfluent speech.


In this work, we offer a novel perspective in enhancing disfluency detection by incorporating visual cues. The first challenge in developing audio-visual multimodal frameworks for disfluency detection is the lack of paired annotated audio-visual disfluency datasets. Second, due to various practical issues such as sensor malfunctions, resource limitations, environment constraints (poor lighting, obstructions, etc.), and privacy concerns, it is common to have missing modalities, particularly video, during inference. 
Finally, unlike mainstream speech enhancement tasks, the paradigms of audio-visual multimodal learning may not seamlessly extend to the assessment of paralinguistic tasks in atypical speech. Hence, designing multimodal learning frameworks tailored to disfluent speech, resilient to modality-missingness, is crucial to improving disfluency detection.


\smallskip\noindent\textbf{Our Approach.} To design a novel multimodal fusion framework resilient to arbitrary missingness of video modality, we first curate a custom audio-visual dataset leveraging meta-data~\cite{lea2021sep} and raw datasets~\cite{ratner2018fluency} from past works. Our architecture features a weight-sharing encoder for both modalities, allowing it to operate on samples where one of the modalities is missing during training or inference, as shown in Figure~\ref{fig:overall_system_diag}. (I). We utilize a temporal decimator to project the densely sampled audio input's feature embedding from a pretrained foundation model to the same vector space as the features extracted from the full facial region of the speaker in video samples. Subsequently, both modalities share a learnable encoder, which learns the corresponding temporal context. The encoder then collapses it before incorporating the dynamic scaling factors for the modalities and feeding to a classifier head. Our empirical results show that commonly-used strategies like concatenating features in a lower dimension for fusion and using cropped lip regions as typically used for speech recognition tasks~\cite{ma2023auto,sadeghi2020audio} do not work very well in this case. Hence, we also demonstrate other fusion strategies when the availability of both modalities is always guaranteed, as shown in Figure~\ref{fig:overall_system_diag}. (II, III).

In summary, our contributions include: (1) a 3.3-hour paired and annotated audio-visual disfluency dataset with benchmarking splits (Section~\ref{subsec:dataset}), (2) a novel fusion technique using weight-sharing encoders for disfluency learning despite missing video modalities (Section~\ref{subsec:unifmodel}), (3) alternative fusion strategies for end-to-end disfluency training without concern for missing modalities (Section~\ref{subsec:unifmodel}), and (4) comprehensive evaluation against unimodal and multimodal baselines, demonstrating the superiority of our audio-visual learning for disfluency detection (Section~\ref{sec:results}). Our open-source code is available\footnote{\url{https://github.com/payalmohapatra/Multimodal-Speech-Disfluency.git}}.


\smallskip\noindent\textbf{Related Works.}
\textit{Emotion understanding} is another paralinguistic task where multimodal learning is conducted beyond audio modality, through visual cues, physiological signals, hand gestures, text, etc.~\cite{dadebayev2022eeg,zhang2023dual},  utilizing extensive datasets for modality-dependent fusion~\cite{ghosh2022novel}, cross-modal interaction structures~\cite{chen2021multimodal}, and collaborative learning~\cite{zhang2023dual}. General \textit{multimodal learning} for addressing modality-missingness applies reconstruction in latent space~\cite{ma2021smil}, statistical imputation~\cite{tran2017missing}, encoding missingness~\cite{mohapatra2023person}, or reinforcement learning-based policy design to weigh incomplete input vectors~\cite{gao2022gradient} dynamically. In contrast, our work introduces a unified encoder scheme featuring weight-sharing and additive fusion for a small-scale end-to-end trainable disfluency detection task.


\section{Approach}

\subsection{Dataset Curation}\label{subsec:dataset}



The lack of multimodal disfluency detection frameworks from an audio-visual perspective is largely due to scarce public paired and annotated audio-visual datasets. We take efforts to curate an audio-visual dataset by leveraging meta-data and open-source databases from past works~\cite{lea2021sep,ratner2018fluency} and adhere to their taxonomic recommendations. FluencyBank contains video recordings from 32 individuals with disfluent speech in the English language. Lea et. al~\cite{lea2021sep} released an annotated dataset of 3-second audio segments for the FluencyBank database containing 4,144 audio clips. They recruit 3 annotators to systematically label each audio segment as one of the five disfluencies -- Blocks (Bl), Word repetition (WP), Sound repetition (SnD), Interjection (Intrj), and Prolongation (Pro) or as fluent speech. One of the challenges with this annotation schema is the inter-annotator disagreement and past works have highlighted its detrimental impact on performance~\cite{mohapatra2022speech}. To address this, we conduct a distillation of the dataset by only retaining samples with a majority vote similar to our previous work~\cite{mohapatra2022speech} and constructing a corpus for FluencyBank containing 4,004 annotated audio clips. Next, we leverage the natural temporal alignment of audio and video databases from FluencyBank and segment the available videos based on the meta-data for the start and end times of a continuous recording from the audio annotations to construct a paired multimodal dataset for FluencyBank. Note that although the FluencyBank corpus originally also offered text transcriptions, past works~\cite{romana2023toward} have highlighted its temporal misalignment and inaccuracies in capturing the frame-level disfluencies to a degree that can be leveraged effectively for overall disfluency categorization. Hence, we focus only on the audio-visual multimodal dataset in this work. All the audio data are sampled at 44.1kHz and the video data at 30 frames per second. The summary of the amount of data available for each type of speech disfluency and their descriptions are given in Table~\ref{tab:dataset}. To facilitate benchmarking for multimodal speech disfluency research, we release this curated audio-visual annotated dataset of approximately 3.3 hours along with the training and evaluation splits for reproducibility\footnotemark[\value{footnote}].

\begin{table}[]
\small
\caption{Audio-Visual dataset from the FluencyBank corpus.}
\vspace{-10pt}
\setlength{\tabcolsep}{2pt}
\label{tab:my-table}
\begin{tabular}{llc}
\hline
\multicolumn{1}{l}{\textbf{Category}} & \multicolumn{1}{l}{\textbf{Description}} & \textbf{Size} \\ \hline
Blocks(Bl) & Long unnatural pauses & 469 \\
Word   Repetition(WP) & Repetition of any word & 337 \\
Sound   Repetition(SnD) & Intra-word phoneme repetition & 428 \\
Interjection(Intrj) & Filler words or non-words & 701 \\
Prolongation(Pro) & Extended sounds within a word & 407 \\
NoStutteredWords & Fluent speech & 1660 \\ \hline
\end{tabular}
\vspace{-20pt}
\label{tab:dataset}
\end{table}

\subsection{Preprocessing}\label{subsec:prep}

\noindent\textbf{Audio Embeddings.}
Past works~\cite{mohapatra2022speech, bayerl22_interspeech, mohapatra2023effect} consistently highlight the effectiveness of features extracted from large-scale pre-trained networks over traditional acoustic features. We leverage large-scale foundation models trained on 1000s of hours of typical speech data to extract meaningful features. In this case, we utilize the base architecture of wav2vec 2.0~\cite{baevski2020wav2vec} to support a small-scale end-to-end training with limited labeled data. The 3-second audio data is resampled to 16kHz and fed to 12 frozen layers of transformers in wav2vec 2.0. In Section~\ref{subsec::results} we also show that one can easily replace this extractor as suitable for their task. The audio input, $\mathrm{x_a} \in \mathbb{R}^{1 \times T_a}$ where $T_a$ is the temporal length of each sample, is transformed to $\mathrm{w_a} \in \mathbb{R}^{C_a \times F_a \times T_a}$. Audio signal is sampled at a much higher rate than video modality. To facilitate projection to a common latent space eventually, we implement an embedding decimator, $\mathcal{F}_{decim} :\mathbb{R}^{C_a \times F_a \times T_a} \xrightarrow{} \mathbb{R}^{C_a \times F_a \times T_{decim}}$, where $T_{decim} < T_a$ using 1D convolution (CNN) layers along the temporal axis. This is followed by an audio-feature encoder, $\mathcal{F}_{audio}$ designed using two layers of 2D CNNs with max-pooling to further summarize the embeddings to a common latent space, $\mathbb{R}^{F \times T}$.

\smallskip\noindent\textbf{Video Embeddings.} The video input $\mathrm{x_v} \in \mathbb{R}^{ 3 \times P \times P \times T_v}$, where $P$ is the pixel resolution of the image stream and $T_v$ is the total number of frames in the video segment. To extract meaningful video features from the full facial region, we leverage Ma et. al's~\cite{ma2022visual} preprocessing pipeline for automatic speech recognition, which is well-established for conventional end-to-end visual speech recognition tasks. This pipeline consists of 3D CNNs with a ResNet-18 encoder which transforms $\mathrm{x_v}$ to $\mathrm{w_v} \in \mathbb{R}^{C_v \times F_v \times T_v}$. After this stage a vision encoder, $\mathcal{F}_{video}$, symmetric to $\mathcal{F}_{audio}$ is employed for feature summarization.

\begin{table*}[]
\caption{Comparison of three variants of our multimodal approach (DAV-early, DAV-late, DAV-unified) and four unimodal and multimodal baselines on balanced accuracy and F1-score across five disfluency tasks. 
The best results are highlighted in \textbf{bold}.}
\vspace{-10pt}
\setlength{\tabcolsep}{1.7pt}
\label{tab:result_summary}
\begin{tabular}{l cc cc cc cc cc c}
\hline
\multirow{2}{*}{Task} &
  \multicolumn{2}{c|}{Blocks} &
  \multicolumn{2}{c|}{Word Repetition} &
  \multicolumn{2}{c|}{Sound Repetition} &
  \multicolumn{2}{c|}{Interjection} &
  \multicolumn{2}{c|}{Prolongation} &
  Avg. \\ 
                  & BA                         & \multicolumn{1}{c|}{F1}                        & BA                        & \multicolumn{1}{c|}{F1}                       & BA                         & \multicolumn{1}{c|}{F1}                           & BA                        & \multicolumn{1}{c|}{F1}                        & BA                        & \multicolumn{1}{c|}{F1}                        & Acc.  \\ 
                  \hline
                  \hline
Audio-Only        & $0.64_{\pm0.01}$           & \multicolumn{1}{c|}{$0.44_{\pm0.01}$}          & $0.62_{\pm0.00}$          & \multicolumn{1}{c|}{$0.35_{\pm0.01}$}         & $0.65_{\pm0.02}$           & \multicolumn{1}{c|}{$0.44_{\pm0.02}$}             & $0.62_{\pm0.02}$          & \multicolumn{1}{c|}{$0.49_{\pm0.03}$}          & $0.66_{\pm0.02}$          & \multicolumn{1}{c|}{$0.44_{\pm0.04}$}          & 0.64 \\
Video-Only        & $0.71_{\pm0.01}$           & \multicolumn{1}{c|}{$0.53_{\pm0.02}$}          & $0.75_{\pm0.02}$          & \multicolumn{1}{c|}{$0.54_{\pm0.01}$}         & $\pmb{0.75_{\pm0.03}}$     & \multicolumn{1}{c|}{$\pmb{0.58_{\pm0.010}}$}      & $\pmb{0.79_{\pm0.01}}$    & \multicolumn{1}{c|}{$\pmb{0.71_{\pm0.01}}$}    & $0.67_{\pm0.01}$          & \multicolumn{1}{c|}{$0.45_{\pm0.01}$}          & 0.73 \\
AT-Dsflnt~\cite{at-disflnt}         & NA                         & \multicolumn{1}{c|}{NA}                        & $0.57$                    & \multicolumn{1}{c|}{$0.71$}                   & $0.37$                     & \multicolumn{1}{c|}{$0.55$}                       & $0.77$                    & \multicolumn{1}{c|}{$0.76$}                    & NA                        & \multicolumn{1}{c|}{NA}                        & 0.57  \\
Auto-AVSR~\cite{ma2023auto}         & $0.57_{\pm0.04}$           & \multicolumn{1}{c|}{$0.33_{\pm0.09}$}          & $0.56_{\pm0.02}$          & \multicolumn{1}{c|}{$0.54_{\pm0.01}$}         & $0.55_{\pm0.02}$           & \multicolumn{1}{c|}{$0.3 _{\pm0.06}$}             & $0.57_{\pm0.02}$          & \multicolumn{1}{c|}{$0.43_{\pm0.03}$}          & $0.55_{\pm0.02}$          & \multicolumn{1}{c|}{$0.31_{\pm0.01}$}          & 0.56  \\ 
\midrule
DAV-early         & $0.70_{\pm0.04}$           & \multicolumn{1}{c|}{$0.53_{\pm0.04}$}          & $\pmb{0.77_{\pm0.02}}$    & \multicolumn{1}{c|}{$\pmb{0.6_{\pm0.04}}$}    & $0.73_{\pm0.02}$           & \multicolumn{1}{c|}{$0.56_{\pm0.02}$}             & $0.8 _{\pm0.01}$          & \multicolumn{1}{c|}{$0.72_{\pm0.01}$}          & $0.68_{\pm0.01}$          & \multicolumn{1}{c|}{$0.47_{\pm0.01}$}          & 0.73 \\
DAV-late          & $0.71_{\pm0.03}$           & \multicolumn{1}{c|}{$0.54_{\pm0.02}$}          & $0.71_{\pm0.04}$          & \multicolumn{1}{c|}{$0.55_{\pm0.05}$}         & $0.70_{\pm0.02}$           & \multicolumn{1}{c|}{$0.54_{\pm0.03}$}             & $\pmb{0.79_{\pm0.01}}$    & \multicolumn{1}{c|}{$\pmb{0.71_{\pm0.01}}$}    & $0.67_{\pm0.01}$          & \multicolumn{1}{c|}{$0.46_{\pm0.04}$}          & 0.72 \\
DAV-unified       & $\pmb{0.71_{\pm0.03}}$     & \multicolumn{1}{c|}{$\pmb{0.55_{\pm0.04}}$}    & $0.75_{\pm0.03}$          & \multicolumn{1}{c|}{$0.58_{\pm0.04}$}         & $\pmb{0.75_{\pm0.01}}$     & \multicolumn{1}{c|}{$\pmb{0.58_{\pm0.02}}$}       & $0.79_{\pm0.02}$          & \multicolumn{1}{c|}{$0.70_{\pm0.03}$}          & $\pmb{0.69_{\pm0.02}}$    & \multicolumn{1}{c|}{$\pmb{0.51_{\pm0.01}}$}    & \textbf{0.74} \\ \hline
\end{tabular}
\vspace{-15pt}
\end{table*}

\subsection{Unified Modality Fusion Network} \label{subsec:unifmodel}

Different from most multimodal frameworks, we propose a novel weight-sharing encoder $\mathcal{G}$, which is common to multiple modalities. Through $\mathcal{F}_{audio}$ and $\mathcal{F}_{video}$ encoders, both modalities are projected to a common latent space. Their respective outputs are normalized and added with a positional encoding using different frequencies of sinusoids similar to~\cite{vaswani2017attention}. $\mathcal{G}$ is achieved using a multi-head (16 heads in our case) attention transformer encoder layer~\cite{vaswani2017attention} realized using three input streams $\mathrm{Q, K, V} \in \mathbb{R}^{T \times F}$ from learnable networks to compute the representative attention, $\mathrm{A} \in \mathbb{R}^{T \times F} $, such that,
$$
\mathrm{S} = \mathrm{Q.K^T} \in \mathbb{R}^{T \times T},  \mathrm{P} = \texttt{softmax}({\mathrm{S}}) , 
\mathrm{A} = \mathrm{PV} \in \mathbb{R}^{T \times F}.
$$
Following the multi-headed attention, a temporal mean pooling $\mathcal{F}_{pool}$ is employed, where $\mathcal{F}_{pool}(\mathrm{A}):\mathbb{R}^{T \times F} \xrightarrow[]{} \mathbb{R}^{1 \times F}$. The intuition for this design is that the relevant temporal dependencies are learned by $\mathcal{G}$ and now it is important to focus on the semantic differentiability of the features. The common latent embedding dimensions for both modalities facilitate the design of this weight-sharing encoder: $
\mathrm{r_a}, \mathrm{r_v} = \mathcal{G}(\mathcal{F}_{audio}(\mathrm{w_a})), \mathcal{G}(\mathcal{F}_{video}(\mathrm{w_v})),
$
where $\mathrm{r_a}, \mathrm{r_v}$ are the results of individual forward passes of both modalities through $\mathcal{G}$. This design is tolerant of the missingness of one of the modalities due to this shared-weights-based fusion. It can support the partial availability of modalities during both the training and inference stages. To learn the dynamic importance of the feature maps from these modalities, we incorporate learnable scalars $\mathrm{c_a}, \mathrm{c_v} \in \mathbb{R}^{F}$, such that  $\mathrm{r} = \mathrm{c_a} \cdot \mathrm{r_a} + \mathrm{c_v} \cdot \mathrm{r_v}$. The overall design of this missingness-resilient audio-visual disfluency detection through unified fusion (\textbf{DAV-unified}) network is shown in Figure~\ref{fig:overall_system_diag}. (I). For scenarios where both modalities are guaranteed to be available, we also propose two variants of multimodal learning framework, early fusion (as shown in Figure~\ref{fig:overall_system_diag}. (II)) and late fusion (Figure~\ref{fig:overall_system_diag}. (III)).

\smallskip\noindent\textbf{Early Fusion (DAV-early).}  We empirically found that a mere concatenation of features from the two domains, $\mathrm{w_a}$ and $\mathrm{w_v}$, does not perform well. So we propose a similar design as the unified fusion network, by adding the projections of $\mathrm{w_a}$ and $\mathrm{w_v}$ and then feeding the result's normalized version to the feature encoder with positional encoding. It is designed similarly to $\mathcal{G}$.

\smallskip\noindent\textbf{Late Fusion (DAV-late).} In this case, most of the initial pipeline of data transformation is identical to the unified fusion network. However, instead of a shared encoder, it contains two dedicated encoders $\mathcal{G}_a$ and $\mathcal{G}_v$ to learn modality-specific weights. The resulting embeddings from these two encoders are then added to fuse deeper in the network.
    
\smallskip\noindent\textbf{Modality Dropout Augmentation.} Often in disfluency applications, audio is the primary modality and video data is missing at times. To support such scenarios during inference and training, we adopt an augmentation strategy of dropping the video data randomly during training to allow the model to learn to switch states when one of the modalities is missing. Following the Bernoulli distribution, we sample a binary masking variable $\mathrm{m}$, for the mini-batch of size $B$. $m_i \sim p^m_i*(1-p)^{(1-m_i)} \forall i \in (0,B-1)$, where $p$ is the probability of dropping set to 0.5 in this case. Although we apply modality dropout to only the video domain, this is a versatile augmentation that can be extended to all modalities as suitable. $\mathrm{x_v}$ is dropped if $\mathrm{m_i}$ is 1.

\subsection{Classifier Head} 

The output after the fusion (in case of early fusion schema, the output of $\mathcal{G}$), $\mathrm{r}$, is given to the classifier head, $\mathcal{C}$, which is realized using a few fully-connected layers. Note that incorporating $\mathcal{F}_{pool}$ also significantly helps in reducing the number of trainable parameters for $\mathcal{C}$. To achieve semantic distinction between the fluent and disfluent speech segments, we optimize the function
$
    \mathcal{L}_\mathrm{CE} = \frac{1}{B}\sum_{i=1}^{B} \mathrm{y}_i \log{\mathcal{C}(\mathrm{r})},
$
where $B$ is the size of a batch in the mini-batch training and $\mathrm{y}_i$ is the one-hot form of the label $y_i$. As a side note, we explored a joint training approach using a cosine similarity loss function, treating the shared encoders as a siamese  network~\cite{mohapatra2023efficient} with $\mathrm{r_a}$ and $\mathrm{r_v}$ along with $\mathcal{L}_\mathrm{CE}$, but did not find significant gains.
\subsection{Training Setup}
All experiments are performed on an Ubuntu OS server equipped with NVIDIA TITAN RTX GPU cards using the PyTorch framework. During model training, we use Adam optimizer with a learning rate from 1e-6 to 1e-4, batch size of 512, and the maximum number of epochs is set to 500 based on the suitability of each setting. Since our class distribution is imbalanced, we used a custom sampler to re-weight the samples per class in every minibatch. We tune these optimization-related hyperparameters for each setting and save the best model checkpoint using early exit based on the minimum value of $\mathcal{L}_\mathrm{CE}$ achieved on 10\% of training data held out for validation.

\section{Results}

\subsection{Experimental Setup}\label{sec:results}
We evaluate the effectiveness of our multimodal disfluency detection approach, including variants \textbf{DAV-united}, \textbf{DAV-early} and \textbf{DAV-late},  against state-of-the-art multimodal and unimodal disfluency, and generic speech recognition methods. We use metrics
$
\text{F1}=\frac{2 \cdot \text{TP}}{\text{TP} + \frac{1}{2}(\text{FP} + \text{FN})}$ and $ \text{Balanced Accuracy (BA)}=\frac{1}{2} \left( \frac{\text{TP}}{\text{TP} + \text{FN}} + \frac{\text{TN}}{\text{TN} + \text{FP}} \right),
$
where TP is true positive, FP is false positive, TN is true negative and FN is false positive, to fairly assess imbalanced datasets. We hold out 20\% of the data for each class for evaluation and provide the exact train test splits for the curated dataset. Every experiment is carried out with 3 different seeds (123, 456, 789) and the mean and standard deviations are reported. We report performance gains as the absolute difference from the baseline. We conduct experiments for five different disfluency-detection tasks and maintain uniform feature extraction pipelines for audio and video inputs across all models unless specifically required by the model. Specifically, the baselines include the following.

\begin{figure}[!b]
\centering
\vspace{-12pt}
\includegraphics[width=0.96\linewidth]{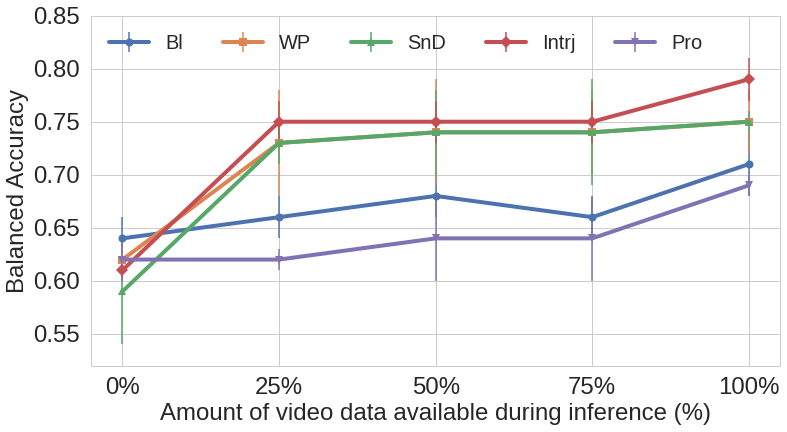}
\vspace{-9pt}
\caption{Performance of our DAV-unified approach on varying availability of visual modality during inference.}
\vspace{-12pt}
\label{fig:sensitivity}
\end{figure}
\smallskip\noindent\textbf{Unimodal Models.} In the \textbf{Audio-Only} model, we use the wav2vec 2.0 BASE feature extractor with 2D CNN layers, a transformer-based encoder, and a classifier head (upper branch in Figure~\ref{fig:overall_system_diag}). For the \textbf{Video-Only} model, we follow a comparable embedding extraction pipeline as outlined in Section~\ref{subsec:prep}, employing a transformer-based encoder and a classifier head (lower branch in Figure~\ref{fig:overall_system_diag}).

\smallskip\noindent\textbf{AT-Dsflnt~\cite{at-disflnt}.}  
We leverage a state-of-the-art audio-text-based multimodal disfluency detector (AT-Dsflnt)~\cite{at-disflnt} with frozen layers, which uses untranscribed audio input, generates text through a custom automatic speech recognition pipeline, and models a multimodal input space for subsequent analysis. Notably, AT-Dsflnt employs a distinct training dataset and disfluency taxonomy. We establish correspondence for three disfluency tasks: filled pauses as interjections, repetitions and word repetitions, and partial words and repetitions as sound repetition; and present results while acknowledging differences in training data and taxonomy.

\smallskip\noindent\textbf{Auto-AVSR~\cite{ma2023auto}.} We leverage a lip-reading-focused audio-visual speech classification state-of-the-art model, Auto-AVSR~\cite{ma2023auto}, for disfluency detection applications. We follow the necessary preprocessing steps for Auto-AVSR by cropping the lip-region of the videos and conducting end-to-end training for the five disfluency tasks.

\subsection{Effectiveness of Audio-Visual Multimodal Learning}\label{subsec::results}
Table~\ref{tab:result_summary} summarizes the performance of four baseline models and the three variants of our multimodal disfluency detection models. We observe that \textbf{our approach boosts the average accuracy of the disfluency detection across all tasks by 10\% and up to 17\% in some cases}. It is evident that visual data significantly enhances the accuracy of disfluency detection. In fact, in two of the cases, the Video-only model can perform as well as the audio-visual multimodal frameworks. This can be attributed to the paralinguistic nature of speech disfluency~\cite{perkins1991theory}, which presents itself distinctly through facial gestures allowing the model to learn semantic disfluency properties accurately.

\smallskip\noindent\textbf{Resilience to missing video modality.}
To demonstrate the resilience of our approach to the missing of video modality, we conduct inference on test sets with varying amount of available video data as shown in Figure~\ref{fig:sensitivity}. Even with 50\% of the visual data missing, our \textbf{DAV-unified offers 7\% average boost in performance} over the unimodal acoustic model.


\begin{figure}[]
\raggedright
\begin{minipage}{0.23\textwidth}
\raggedright
\includegraphics[width=\textwidth]{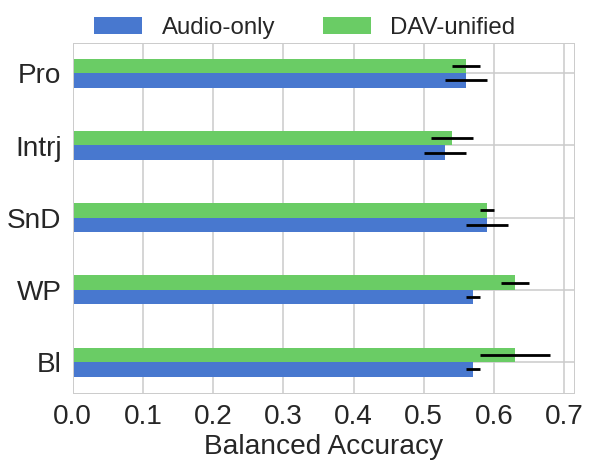}
\vspace{-18pt}
\caption{Performance on zero-shot transfer to a different acoustic dataset (SEP28k).}
\vspace{-18pt}
\label{fig:sep_28k}
\end{minipage}%
\begin{minipage}{0.25\textwidth}

\captionof{table}{Balanced Accuracy (BA) for different types of audio and video features for Blocks.}
\vspace{-10pt}
\centering
\resizebox{1.\textwidth}{!}{
\setlength{\tabcolsep}{0.85mm}{
\begin{tabular}{lcl}
\hline
Features                   & Model       & BA \\ \hline
\hline
Full Face    & Video-only  &  $0.71_{\pm0.01}$   \\ \cline{2-3}
+ wav2vec2.0 & DAV-unified &  $0.71_{\pm0.03}$   \\ \hline
Lip ROI      & Video-only  &  $0.57_{\pm0.02}$   \\ \cline{2-3}
+ wav2vec2.0 & DAV-unified &  $0.59_{\pm0.03}$   \\ \hline
Hubert       & Audio-only  &  $0.58_{\pm0.01}$   \\ \cline{2-3}
+ Full Face  & DAV-unified &  $0.68_{\pm0.02}$   \\ \cline{2-3}
+ Lip ROI    & DAV-unified &  $0.57_{\pm0.04}$   \\ \hline
\end{tabular}}}
\vspace{-20pt}
\label{tab:sensitivity}
\end{minipage}
\end{figure}

\smallskip\noindent\textbf{Additional Insights.}
We conduct a zero-shot cross-dataset transfer, evaluating models trained on the FluencyBank on SEP28k, consisting solely of audio samples. In the absence of video input for this task, our DAV-unified model matches the performance of an Audio-Only model, with up to a 6\% boost in balanced accuracy, highlighting its generalization ability.

We examine the impact of various audio-visual features on speech disfluency tasks in Table~\ref{tab:sensitivity}. Visual features are extracted from both the full face and the cropped lip region of interest (ROI)~\cite{ma2022visual}, and audio features are from the base versions of Hubert and wav2vec 2.0. Unlike mainstream audio-visual speech recognition trends, focusing on the lip region diminishes performance, with an average degradation of 14\% and 11.5\% in Video-Only and DAV-unified, respectively, compared to full facial features. Reasons for this drop could be the 18\% less paired data available for training the lip ROI cases due to the inability to detect landmarks and video quality issues for some samples. Moreover, the paralinguistic nature of disfluency implies potential advantages derived from a comprehensive facial context. While these findings are intriguing, it is essential to recognize that larger databases and variations in pretrained networks for visual features could impact and shape this observation.

\section{Conclusion}

We present a novel audio-visual multimodal learning approach for speech disfluency detection,  curating a FluencyBank dataset and developing a unified weight-sharing multimodal design that is resilient to missing video modality. Our findings highlight the substantial performance gains from video, even when present partially in a multimodal framework, 
surpassing the commonly used acoustic-only modality for disfluency detection. 

\clearpage
\bibliographystyle{IEEEtran}
\bibliography{mybib}

\end{document}